\documentclass[10pt,twocolumn,letterpaper]{article}

\usepackage[pagenumbers]{wacv} 

\usepackage{graphicx}
\usepackage{physics}
\usepackage{booktabs}

\usepackage{amsfonts}
\usepackage{subcaption}
\usepackage{multirow, enumitem, float, wrapfig}
\usepackage{amsmath, amssymb, mathtools, bm}
\usepackage{tabulary,overpic,xcolor}
\usepackage{mathrsfs}
\usepackage[accsupp]{axessibility}  
\usepackage{pifont}
\newcommand{\cmark}{\ding{51}}%
\newcommand{\xmark}{\ding{55}}%

\usepackage[pagebackref,breaklinks,colorlinks]{hyperref}

\usepackage[capitalize]{cleveref}
\crefname{section}{Sec.}{Secs.}
\Crefname{section}{Section}{Sections}
\Crefname{table}{Table}{Tables}
\crefname{table}{Tab.}{Tabs.}

\def\wacvPaperID{1901} 
\def\confName{WACV}
\def\confYear{2025}

\begin{document}


\title{A Two-Stage Progressive Pre-training using Multi-Modal Contrastive Masked Autoencoders}

\author{Muhammad Abdullah Jamal and Omid Mohareri \\
Intuitive Surgical Inc.}
\maketitle

\begin{abstract}
In this paper, we propose a new progressive pre-training method for image understanding tasks which leverages RGB-D datasets. The method utilizes Multi-Modal Contrastive Masked Autoencoder and Denoising techniques. Our proposed approach consists of two stages. In the first stage, we pre-train the model using contrastive learning to learn cross-modal representations. In the second stage, we further pre-train the model using masked autoencoding and denoising/noise prediction used in diffusion models. Masked autoencoding focuses on reconstructing the missing patches in the input modality using local spatial correlations, while denoising learns high frequency components of the input data. Moreover, it incorporates global distillation in the second stage by leveraging the knowledge acquired in stage one. Our approach is scalable, robust and suitable for pre-training RGB-D datasets. Extensive experiments on multiple datasets such as ScanNet, NYUv2 and SUN RGB-D show the efficacy and superior performance of our approach. Specifically, we show an improvement of +1.3\% mIoU against Mask3D on ScanNet semantic segmentation. We further demonstrate the effectiveness of our approach in low-data regime by evaluating it for semantic segmentation task against the state-of-the-art methods.
\end{abstract}


\section{Introduction}
\label{sec:intro}

Self-supervised learning (SSL) has paved the way for improving performance in various computer vision tasks such as object recognition~\cite{mae,simmm,beit}, semantic segmentation~\cite{mask3d,gao2022convmae,li2022semmae} and depth estimation~\cite{guizilini2022multiframe} by leveraging extensive unlabeled data and large models such as Vision Transformers (ViT)~\cite{vit,li2022mvitv2}. There are two self-supervised frameworks commonly used to pre-train large vision models. The first family of SSL is Masked Image Modeling (MIM) which is inspired by the recent success of masking in the NLP domain~\cite{BERT}. The seminal work in MIM is Masked Autoencoders (MAE)~\cite{mae} which masks out high number of patches in the input before passing them into a Transformer encoder. Then, a lightweight decoder is used to reconstruct the masked patches in the input. Such reconstruction task provides the model with understanding of the data beyond low-level image statistics. It has achieved state-of-art performance in a variety of image recognition tasks.

Recently, Masked autoencoders have been extended to pre-train transformers using multi-modal data for tasks such as video-text retrieval~\cite{li2023harvest,wang2023unified}, image-text retrieval~\cite{clip_} and video-audio-text understanding~\cite{akbari2021vatt}. However, utilizing MAEs for dense prediction tasks like semantic segmentation and depth estimation requires understanding beyond image level statistics. With the availability of RGB-D or Time-of-Flight (ToF) cameras, various approaches~\cite{multimae,mask3d,girdhar2022omnivore,du2021cross} have been developed to improve the dense prediction tasks using multi-modal datasets~\cite{Armeni3D,Matterport3D,scannet,nyuv2,sunrgbd}. A few methods~\cite{multimae,mask3d} also leverage RGB-D datasets to pre-train ViTs backbones under MAE paradigm. One such work is MultiMAE~\cite{multimae} which pre-trains the vision transformers using RGB, depth and segmentation masks in a multi-task MAE fashion. However, MultiMAE requires semantic segmentation labels during pre-training and access to multiple modalities during fine-tuning for downstream tasks. This method learns cross-modal relationships using a cross-attention module but under the paradigm of MAE to help reconstruct the masked patches in each modality. Another work in this line is Mask3D~\cite{mask3d} which embeds 3D priors in ViT backbones to learn structural and geometric priors by reconstructing the masked patches in the depth input using RGB-D data. However, it doesn't learn cross-modal relationships to capture more local context beyond 3D priors. 

The second family, contrastive learning~\cite{simclr, mocov2,Mocov3,byol} pre-trains a model by maximizing the similarity between different augmentations of an image while minimizing the similarity to the other data samples. It essentially learns discriminative representations by enforcing invariance to data augmentations. Some prior works~\cite{chen20224dcontrast,xie2020pointcontrast,zhang2021selfsupervised,hou2021exploring} have explored contrastive learning for high-level scene understanding and low-level point matching~\cite{banani2021bootstrap,zhang2023pcrcg}.

Motivated by this, we believe contrastive learning and Masked autoencoders can be complementary to each other as both can provide unique data representations by extracting different discriminative features from RGB-D datasets. We have seen recent works that combine instance-level contrastive learning and masked autoencoders and leverage multi-modal data in one single framework for image recognition~\cite{cmae}, video action recognition~\cite{lu2023cmaev}, video-text~\cite{ma2022simvtp} and audio-video~\cite{cavmae}. However, such methods are not suitable for RGB-D datasets. For example, dense prediction tasks such as semantic segmentation and depth estimation require understanding beyond image-level representation. We hypothesize that the instance-level contrastive learning would not benefit the aforementioned downstream tasks when leveraging RGB-D datasets for pre-training. So, to capture useful discriminative local context features using these modalities, patch-level or pixel-level contrastive learning would be more beneficial. However, it is not easy to combine patch-level or pixel-level contrastive learning with Masked Autoencoders in a single framework because of the masking step. Recently, CoMAE~\cite{comae} proposed a single model self-supervised hybrid pre-training framework for small-scale RGB-D datasets. It incorporates contrastive learning and multi-modal masked autoencoding to pre-train a single encoder to capture cross-modal representations. However, it requires both RGB and depth data during fine-tuning and evaluation, as omitting either results in a significant drop in downstream performance. It is also limited to pre-training with small-scale datasets and fails to distill features learned in stage-1, as stage-2 continues to pre-train the encoder using only the low-level pixel reconstruction task. This might not capture rich and diverse representations, and it is an easier task and can be learned quickly because of stage-1 initialization.


Additionally, recent approaches frequently struggle to capture high-frequency components of the data, which further motivates us to explore alternative learning mechanisms. Notably, recent advancements in diffusion models have achieved remarkable success in image and video generation~\cite{ho2020denoising,ho2022imagen,ho2022video} by leveraging denoising techniques, a strategy that has been absent in previous approaches, prompting us to investigate its potential in our context. Inspired by this, we introduce a noise prediction loss with the masked autoencoder reconstruction objective. To do so, we add a Gaussian noise in the input and pre-train the model to predict the masked patches as well as the noise. We hypothesize that the denoising would encourage the model to extract high-frequency features from the input.

Based on the above challenges, we propose a new strategy based on these self-supervised frameworks to pre-train ViTs using RGB-D datasets. Our pre-training strategy consists of two stages, 1) contrastive learning at patch-level to align RGB and depth modalities to capture local context for cross-modal relationships 2) MIM + Denoising would encourage the encoders to learn discriminative low-frequency and high-frequency features respectively. Besides, we capitalize on the feature distillation by distilling the global embeddings from the first stage, improving upon MIM + Denoising.


Lastly, we pre-train ViTs on ImageNet~\cite{imagenet}, ScanNet~\cite{scannet} and SUN RGB-D~\cite{sunrgbd} with our progressive pre-training and evaluate it on multiple downstream tasks. 

In summary, our main contributions are:
\begin{itemize}
    \item We present a new strategy for pre-training ViTs using RGB-D dataset. Our approach provides different flavours of self-supervised frameworks i.e, contrastive learning, masked autoencoding and feature distillation.

    \item Inspired by the success of denoising in the recent diffusion models, we also propose to add a noise prediction objective to learn different discriminative features.

    \item We empirically demonstrate the effectivness of our approach on multiple datasets such as ScanNet~\cite{scannet}, NYUv2~\cite{nyuv2} and SUN RGB-D~\cite{sunrgbd} for downstream tasks like semantic segmentation and depth estimation. 

    \item Finally, we show the data-efficient nature of our pre-training approach by evaluating it under limited labeled data scenarios.
\end{itemize}


\section{Related Work}


\paragraph{Contrastive learning based pre-training.} SSL consists of designing a pre-text task to learn discriminative representations from unlabeled data. Earlier work on constructing pre-text tasks include colorization~\cite{zhang2016colorful}, jigsaw puzzle~\cite{jigsaw}, and predicting the rotation angle of image~\cite{rotnet}. Currently, the two popular approaches for self-supervised learning are constrastive learning and masked image modeling. Contrastive learning groups similar data samples in the feature space while pushing away the dissimilar ones. SimCLR~\cite{simclr} uses contrastive learning to maximize the similarity between different augmentations of an image. MoCo~\cite{mocov2} uses dictionary as a queue to facilitate contrastive learning. BYOL~\cite{byol} uses two networks termed as online and target, to learn from each other by using augmented views of an image. On the other-hand, there are clustering based unsupervised methods that either use k-means assignments as pseudo-labels~\cite{deepcluster} or enforce similarity between the cluster assignments of different augmented views of the image~\cite{swav}.

\paragraph{Masked Autoencoder based pre-training.} Masked Image Modeling (MIM) has received a lot of attention for pre-training vision models since the advent of Vision Transformers (ViT). The goal of MIM is to learn discriminative features for downstream tasks by reconstructing the missing patches in the input. Masked autoencoding for vision tasks is inspired by the success of methods such as BERT~\cite{BERT} and GPT~\cite{GPT} in NLP domain. Several MIM approaches have been proposed with various reconstruction objectives based on pixels~\cite{SIT_method,vit,iGPT,elnouby2021largescale,mae,simmm,surgmae}, dVAE tokens~\cite{beit,ibot} and features such as histogram of oriented gradients (HOG) or deep neural network embeddings~\cite{maskfeat,data2vec}. Specifically, MAE~\cite{mae} uses an asymmetric encoder-decoder architecture based on ViT to pre-train models using masked autoencoding. It first masks large portions of an image and pass the unmasked patches into encoder followed by a decoder for reconstructing the missing patches. The masking accelerates the pre-training of ViT. Contrastive learning is different from MAE in the sense that MAE learns local spatial dependencies, while contrastive approaches learn representations such as invariance to different augmentations. This steers us to hypothesize that these two SSL frameworks learn complimentary features and therefore we combine them in a multi-stage manner to get the best out of each. Indeed, one can combine them in a single framework~\cite{cmae,cavmae}, but our empirical experiments show poor performance of such methods on RGB-D datasets.

\paragraph{Multi-Modal Learning.} It involves learning from multiple modalities such as images and text~\cite{alayrac2020selfsupervised,kaiser2017model,e2evlp,castrejon2016learning,hu2021unit,chen2020uniter,lxmert}, video and audio~\cite{arandjelović2017look,jaegle2022perceiver,nagrani2022attention,owens2018audiovisual}, video, text and audio~\cite{akbari2021vatt}, and RGB and depth images~\cite{girdhar2022omnivore}. Learning can be performed either using modality-specific encoders or shared ones. In terms of MIM, MultiMAE~\cite{multimae} pre-trains the ViT using RGB, depth and semantic segmentation  while  reconstructing the missing patches in these modalities. It also uses additional modality during the fine-tuning. Mask3D~\cite{mask3d} uses RGB and depth modalities under MIM paradigm to encode the 3D priors by reconstructing the missing patches in the depth input. CoMAE~\cite{comae} proposes a hybrid pre-training framework consists of contrastive and masked autoencoding for small-scale RGB-D datasets.

    
\paragraph{Denoising.} It involves reconstructing clean data given a noisy input. Denoising has recently revolutionized generative models that produce high quality images using diffusion modeling~\cite{dhariwal2021diffusion,ho2020denoising}. It consists of forward Gaussian diffusion process and backward generation process. In the forward process, it adds Gaussian noise to the sample while in reverse process, it learns to generate high quality image from the Gaussian noise. This has been used for generating text-conditioned images~\cite{nichol2022glide,ramesh2022hierarchical,saharia2022photorealistic} and videos~\cite{singer2022makeavideo,ho2022imagen,ho2022video}. The success of denoising hasn't been utilized in representation learning and in this work, we leverage the denoising process to investigate the effectiveness of  representations learned from RGB-D data. 

\paragraph{Knowledge Distillation.} Knowledge distillation~\cite{Gou_2021,hinton2015distillingnn,phuong2021understandingknowledgedistillation} aims to transfer knowledge from one model mainly called teacher to the second model called student. It treats the output of the teacher model as the target prediction for the student model. Some approaches~\cite{hinton2015distillingnn,zhao2022decoupledknowledgedistillation,chen2022simkd} use logits of the teacher model to distill knowledge while some methods~\cite{wang2022semckd,giftkd,ickd,FitNets} leverage the intermediate features for distillation. In this paper, we present the first attempt to incorporate feature distillation for RGB-D pre-training.
\vspace{-5pt}
\section{Method} \label{sec:approach}
Our approach is a simple combination of constrastive learning, masked autoencoding, denoising used in diffusion models and feature distillation, designed in a two-stage pre-training for vision transformers using RGB-D data. Our goal is to learn representations from different learning mechanisms that can complement each other. We first give an overview of our approach and then explain the learning objectives i.e, contrastive, masked image modeling, noise prediction and distillation. Finally, we describe the overall progressive pre-training pipeline.


\subsection{Overview}
\begin{figure*}
    \centering
    \includegraphics[width=0.80\textwidth]{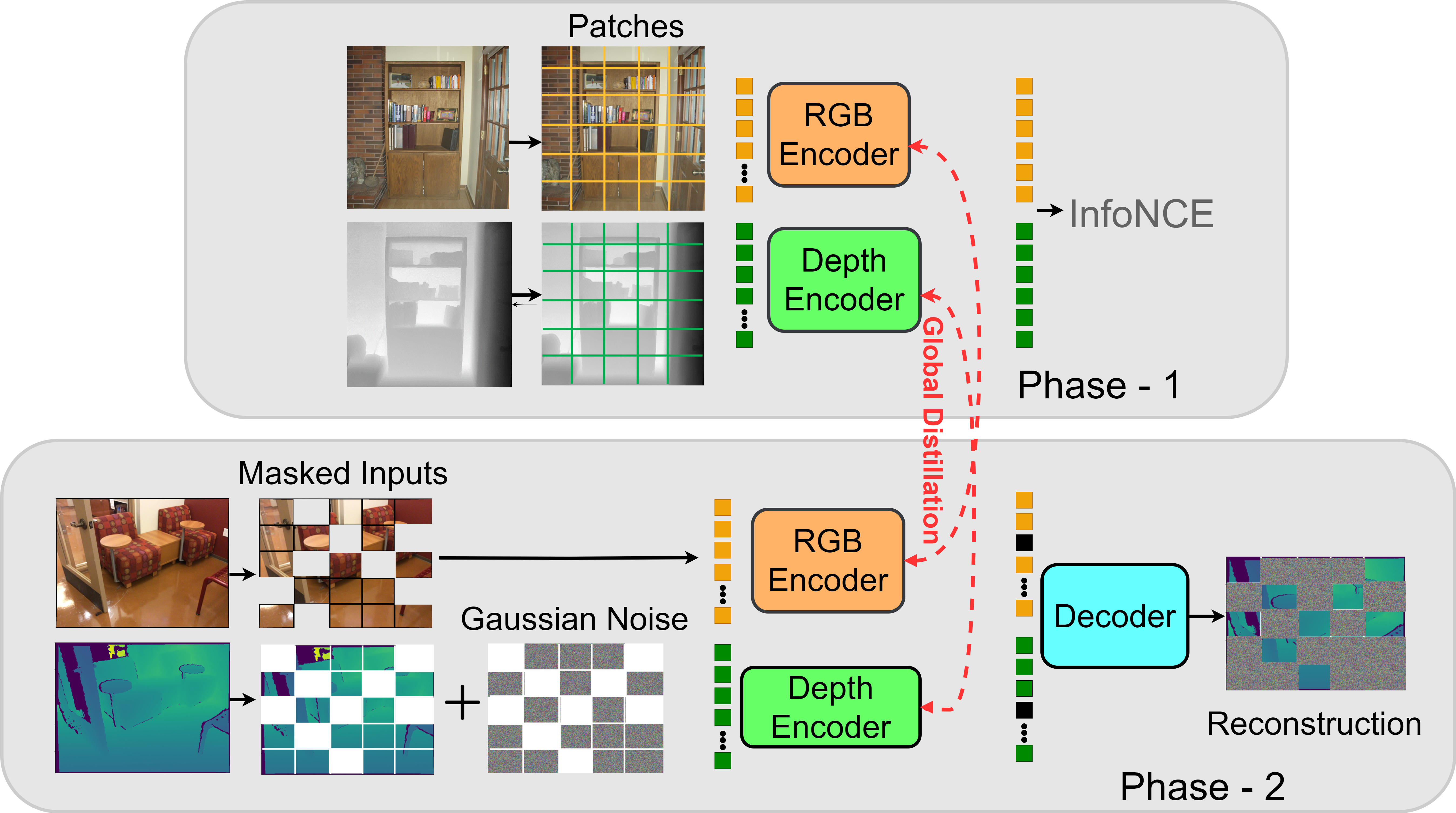}
     \vspace{5pt}
    \caption{\textbf{Overview of our progressive pre-training.} In the first stage, we pre-train the encoders using \textbf{Contrastive learning} to align the RGB and the depth patches. In the second stage, we initialize the modality-specific encoders with the stage-1 weights and pre-train them using \textbf{Multi-modal Masked autoencoding} and \textbf{Denoising} to reconstruct the masked patches in depth input and the noise in the unmasked patches respectively. Moreover, we incorporate \textbf{Feature distillation} to leverage the knowledge acquired in the stage-1.}
    \label{fig:approach}
    \vspace{-5pt}
\end{figure*}

Given a batch of $n$ RGB-D pair $\{\vb{x}^{rgb} , \vb{x}^{depth}\}_{i=1}^n$, we pre-train ViT~\cite{vit} using the combination of self-supervised frameworks. Our pre-training consists of two stages. In the first stage, we pre-train our model to learn cross-modal representations to capture correspondence using constrastive learning at patch level. In the second stage, we first initialize the encoders using the weights learned in the previous stage. Then, we use multi-modal masked autoencoder to learn more fine-grained features beyond low-level image statistics to reconstruct the missing patches in the depth input. To pre-train using masked autoencoding, we first split the RGB-D input in $T= (h/p) \times (w/p)$ patches of size $p \times p$:  $\vb{x}_{i,\text{patch}}^{rgb}, \vb{x}_{i,\text{patch}}^{depth} \in \mathbb{R}^{T \times p \times p \times 3}$ to pass them to the encoders. We use modality-specific encoders in the both stages. We then randomly mask patches in both inputs by keeping a percentage of patches and obtain two masks $\vb{M}_i^{rgb}, \vb{M}_i^{depth}$. Following MAE~\cite{mae}, we pass the unmasked patches to the encoders to obtain the latent representations. The representations are then concatenated with the learnable masked tokens which are essentially the placeholders for these tokens. A lightweight decoder is used to reconstruct the masked patches. Furthermore, we also add denoising to learn the high frequency information of data. Additionally, we utilize global distillation loss to efficiently leverage the knowledge acquired in the first stage. Figure~\ref{fig:approach} shows the overview of our approach.

\subsection{Contrastive learning} \label{sec:contrastive}
To learn the correspondence between RGB and depth modalities, we use contrastive learning to discriminate the instances at patch-level. One can also apply contrastive learning on pixel-level by up-sampling the features to the input resolution. There has been a few works~\cite{liu2021learning,li2022closer} using constrastive learning to learn correspondence between RGB and pointclouds. The depth values can be indirectly related to the pointclouds and projected to 3D pointclouds using camera parameters. Therefore, we can also leverage the same framework for pixel-level of pixel-to-point~\cite{liu2021learning} to learn the correspondence between RGB and depth. This is left for the future work. Specifically, for a given RGB $\vb{x}_{i}^{rgb}$ and depth $\vb{x}_{i}^{depth}$, we pass them through the ViT encoders to get the latent representations $\vb{z}^{rgb}_i, \vb{z}_i^{depth}$. We use \texttt{Conv2d} as patch projection layer. Following~\cite{oord2019representation}, we use InfoNCE loss to pre-train the encoder with a fixed temperature $\tau$:

\begin{equation}
\mathcal{L}_\mathrm{PNCE} = - \frac{1}{N} \sum_{i=1}^N {\rm log}  \left[ \frac{ {\rm exp} (s_{i,i}/\tau)}{\sum_{k \neq i} {\rm exp} (s_{i,k}/\tau) + {\rm exp} (s_{i,i}/\tau)} \right]
\label{eq:contrastive loss}
\end{equation}
where $s_{i,j} = \|\vb{z}^{rgb}_i\|^T\|\vb{z}_i^{depth}\|$. $\|\vb{z}^{rgb}_i\|$ and $\|\vb{z}_i^{depth}\|$ correspond to the encoder features for RGB and depth respectively for a patch or pixel $i$. In the experiments, we apply the InfoNCE loss between $\vb{z}^{rgb}_i$ , $\vb{z}_i^{depth}$ , $\vb{z}^{depth}_i$ and $\vb{z}_i^{rgb}$, and take the average of the losses. This loss aligns the RGB and depth patch by maximizing the similarity using the latent representations, while minimizing the similarity with other depth patches for a given RGB patch. Indeed, an instance-level contrastive learning can be used but it would fail to capture local discriminative features across modalities as the task becomes much easier by only learning high-level semantics.

\subsection{Masked Autoencoding} \label{sec:mae}
The goal of masked autoencoding is to understand local spatial statistics of the input modality. To further improve the capability of our approach, we propose to pre-train the model using MAE~\cite{mae} as a second learning step. We initialize our modality-specific encoders with the stage-1 weights. Unmasked patches of two modalities are passed to the encoders to get latent embeddings. Embeddings are then concatenated with learnable masked tokens which are then passed to the lightweight decoder. We then apply depth prediction MLP head to reconstruct the missing patches. We also experiment with RGB reconstruction task but we found that empirically it doesn't improve the downstream performance (c.f. Table~\ref{tab:ablation_loss}). Following~\cite{mae,mask3d}, we normalize the target patches and compute the loss only on masked patches. Specifically, we use MSE loss which is given as:

\begin{equation}
    \mathcal{L}_\mathrm{depth} = \frac{1}{n} \sum_{i=1}^n \| \vb{M}_i^{depth} \circ ( \vb{x}^{depth}_i - \hat{\vb{x}}^{depth}_i) \|_2^2
    \vspace{-5pt}
\end{equation}

\begin{figure}
    \centering
    \includegraphics[width=0.48\textwidth]{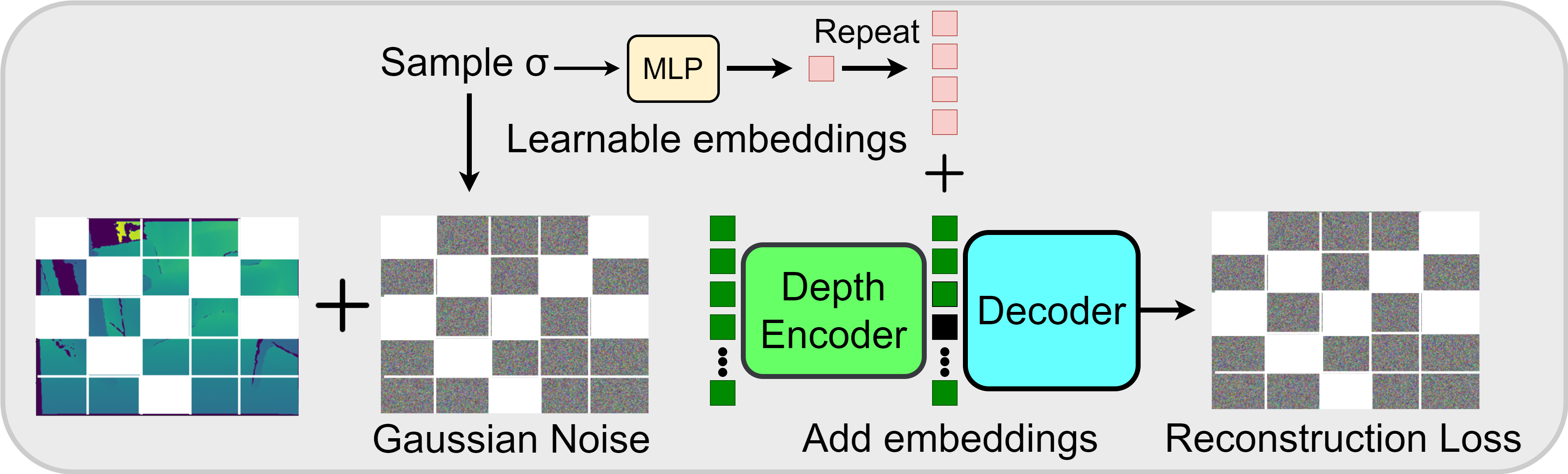}
    \caption{\textbf{Denoising:} We first add Gaussian noise to the unmasked patches. We then pass the noise level $\sigma$ through MLP, and add it to encoded tokens before passing them through the decoder for  reconstruction.}
    \label{fig:denoising}
    \vspace{-10pt}
\end{figure}

\subsection{Denoising}
Denoising approaches has been shown to be a catalyst for models generating high quality images and videos~\cite{ho2020denoising,ho2022imagen}. Inspired by them, we investigate whether denoising can also be used in self-supervised learning. During pre-training, we add an independent isotropic Gaussian noise to the depth modality $\vb{x}_i^{depth} \leftarrow \vb{x}_i^{depth} + \sigma_i^{depth} \vb{e}^{depth}_i$ with $ \vb{e}^{depth}_i \sim \mathcal{N}(\vb{0} , I)$ and $\sigma_i^{depth}$ uniformly sampled from an interval $[0, \sigma_\text{max}]$. This noisy input is masked and then passed to the depth encoder as mentioned in section \ref{sec:mae} which follows MAE~\cite{mae}. Before passing the encoded representations to the decoder, we also add the information about noise level $\sigma_i^{depth}$ to help separate the noise from the clean input, which is inspired from the denoising diffusion model~\cite{ho2020denoising}. To differentiate, we treat $\sigma_i^{depth}$ as the positional encoding to the decoder. We first produce a 1D sinusoidal embedding of $\sigma_i^{depth}$ which is then passed to the MLP with ReLU activation to produce a learnable embedding $\vb{p}_i^{depth} \in \mathbb{R}^d$. The dimension of this embedding matches the dimension of the representation from the encoder. We then add this embedding to the encoded embeddings as $(\vb{z}_i^{depth}) \leftarrow (\vb{z}_i^{depth}) + \vb{p}_i^{depth}$ before concatenating it with the RGB encoder representations and learnable masked tokens. The concatenated result is passed to the decoder to produce $\hat{\vb{x}}^{depth}_i$. We use the MSE loss and compute the denoising loss on the unmasked patches as:

\begin{equation}
    \mathcal{L}_\mathrm{denoise} = \frac{1}{n} \sum_{i=1}^n \| (1-\vb{M}_i^{depth}) \circ (\sigma_i^{depth} \vb{e}^{depth}_i - \hat{\vb{x}}^{depth}_i) \|_2^2 
\end{equation}

Denoising would encourage the encoder to extract high frequency components which can be complementary to the MAE pre-training in the second stage. The computation of the denoising doesn't bring any overhead to the pre-training compared to the vanilla MAE, since we use a lightweight MLP and compute the loss on the reconstructed unmasked patches. The decoder also reconstructs the unmasked patches which may go wasted in the vanilla MAE pre-training. We also provide an ablation study (c.f. Table~\ref{tab:ablation_denoising}) on the effect of each component in the denoising stage. Lastly, the figure~\ref{fig:denoising} shows the overall setup of the denoising.

\subsection{Feature Distillation}
To distill the global modality-specific embeddings in the second stage, we use the output of the encoders as the visual embeddings for each modality. We then apply max pooling to output a global embeddings. Afterwards, we align this second-stage global embedding to the global visual embedding from the first-stage model using smooth $\ell_1$ loss which is given as:

\begin{equation}\label{eq:supervise}
    \mathcal{L}_\mathrm{distill} (\mathbf{f_2}, \mathbf{f_1}) = \begin{cases}
\frac{1}{2} (\mathbf{f_2} - \mathbf{f_1})^2/\beta, & | \mathbf{f_2} - \mathbf{f_1} | \leq \beta \\
(|\mathbf{f_2} -\mathbf{f_1}|-\frac{1}{2}\beta), & \text{otherwise}
\end{cases},
\end{equation}

where $\beta$ is set to 2.0; $\mathbf{f_2}$ and $\mathbf{f_1}$ are the max-pooled features from the stage-2 and stage-1 encoders respectively. In practice, we employ the feature distillation for both RGB and depth. So, the final distillation loss is the sum of modality-specific distillation losses.

\subsection{Progressive Pre-training}
We are now ready to present our full algorithm for pre-training ViT using RGB-D datasets. In the first stage, we pre-train the model using contrastive loss as mentioned in section~\ref{sec:contrastive}. 

\begin{equation}
    \mathcal{L}_\mathrm{stage1} =  \mathcal{L}_\mathrm{PNCE}
\end{equation}
In the second stage, we pre-train the model using Masked autoencoding, denoising and feature distillation. Therefore, the overall stage-2 objective can be defined as:
\begin{equation}
    \mathcal{L}_\mathrm{stage2} =  \alpha \mathcal{L}_\mathrm{depth} + \beta \mathcal{L}_\mathrm{denoise} + \gamma \mathcal{L}_\mathrm{distill}
\end{equation}\label{eq:totalloss-stage2}
Where $\alpha$, $\beta$ and $\gamma$ are the weights for each loss function during pre-training. Please see appendix on the details of these hyper-parameters.

\section{Results}
In this section, we show the efficacy of our approach for ViT-B backbone on semantic segmentation, instance segmentation and depth estimation. In the first stage, we pre-train the ViT-B encoders using ImageNet~\cite{imagenet} dataset. We use off the shelf depth estimation model~\cite{AdaBins} to extract the depths. In the second stage, we continue pre-training the encoders on the ScanNet~\cite{scannet} dataset which contains 2.5M RGB-D frames from 1513 training video sequences. We regularly sample every 25$^{th}$ frame without any filtering during pre-training following Mask3D~\cite{mask3d}. To study the effectiveness of our pre-training approach for small scale datasets, we also pre-train the model using SUN RGB-D dataset~\cite{sunrgbd} in the second stage. It contains 10,335 RGB-D images captured from 4 different sensors. We use the official split of train/test for pre-training and evaluation. We also initialize the second stage encoders with the weights learnt in the first stage. Please refer to the appendix for more details on implementations and hyper-parameters.


\begin{table*}[h!]
\centering
\resizebox{0.99\textwidth}{!}{%
\begin{tabular}{c|c|c|c|c|c}
\hline

\hline

\hline
Methods  & Reconstruction task  & Backbone &  Pre-train  & Fine-tune Modality  & mIoU \\
\hline

\hline

\hline
Scratch  & -                                                                       & ViT-B    & \textcolor{gray}{None} & RGB     & 32.6  \\ \hline

Pri3D~\cite{pri3d_}  &                                                                        & ViT-B   & ImageNet+ScanNet  & RGB      & 59.3  \\ \hline

Pri3D~\cite{pri3d_}  & -                                                                       & ResNet-50   &  ImageNet+ScanNet & RGB        & 60.2  \\ \hline

DINO~\cite{dino}  & -                                                                       & ViT-B   & ImageNet+ScanNet  & RGB        & 58.1 \\ \hline

MAE~\cite{mae}  & RGB                                                                      & ViT-B    &  ImageNet& RGB        & 64.8  \\ \hline

MAE~\cite{mae}  & RGB                                                                       & ViT-B      & ImageNet+ScanNet  & RGB       & 64.5 \\ \hline

MultiMAE*~\cite{multimae}  & RGB + Depth                                                                       & ViT-B    & ImageNet+ScanNet&RGB       & 65.1  \\ \hline

Mask3D**~\cite{mask3d}   & Depth                                                                       & ViT-B    & ImageNet+ScanNet       & RGB        & 66.2  \\ \hline
Mask3D~\cite{mask3d} &  RGB + Depth                                                                   & ViT-B    & ImageNet+ScanNet  & RGB & 65.5 \\ \hline

\textbf{Ours}  & Depth & ViT-B    & ImageNet+ScanNet   & RGB        & \textbf{67.5}  \\
\hline

\hline

\hline
MultiMAE~\cite{multimae} & RGB + Depth + Segmentation & ViT-B & ImageNet  & RGB   & 66.4 \\ \hline
\end{tabular}
}%
\vspace{5pt}
\caption {\label{tab:scannet} \textbf{ScanNet Semantic Segmentation}. Our approach outperforms Mask3D and other SOTA approaches that leverage RGB-D data during pre-training. * means MultiMAE was pre-trained using ScanNet by initializing the ViT-B encoder with ImageNet SSL weights. ** means ConvNext decoder is used instead of transformer decoder. }
\end{table*}

\subsection{Semantic Segmentation}
We first present the results for semantic segmentation. Following MultiMAE~\cite{multimae}, we use ConvNext architecture~\cite{convnext} as a segmentation decoder for our approach and the baselines. All the baselines are initialized with ImageNet self-supervised weights~\cite{mae}. For evaluation, we use mean Intersection over Union (mIoU). Table~\ref{tab:scannet} shows the performance of our approach and other baselines on ScanNet 2D semantic segmentation task. For fine-tuning, we follow Mask3D~\cite{mask3d} and sample every 100$^{th}$ frame, resulting in 20,000 training images and 5000 validation images. Our approach achieves 67.5 mIoU compared to 66.2 mIoU for Mask3D and 65.1 mIoU for MultiMAE. For fair comparison, we pre-train MultiMAE with RGB and depth only and use ConvNext decoder for Mask3D. Moreover, compared to MAE pre-trained on ScanNet RGB modality, our approach improves the performance by \textbf{+3.0} mIoU. Furthermore, compared to Pri3D, a multi-view contrastive learning approach, we outperforms it by \textbf{+7.3} mIoU. Table~\ref{tab:sunrgbd} shows the results on SUN RGB-D 2D semantic segmentation. We observe that our method outperforms other state-of-the-art approaches. More notably, we achieve 48.7 mIoU compared to 47.4 mIoU for Mask3D and 47.1 mIoU for MultiMAE. Table~\ref{tab:sunrgbd_acc} shows the comparison of our approach and CoMAE~\cite{comae} on the SUN RGB-D dataset in terms of mean accuracy, as CoMAE did not report mIoU results for this dataset. We observe that our approach outperforms the CoMAE by 1.1\%. These results demonstrate that our second-stage pre-training is effective even on relatively small-scale RGB-D datasets. We also present results on NYUv2 semantic segmentation, where similar observations can be made when comparing our method with the competing approaches. For further details, please refer to the supplementary materials.

\begin{table*}[h!]
\centering

\resizebox{0.99\textwidth}{!}{%
\begin{tabular}{c|c|c|c|c|c}
\hline

\hline

\hline
Methods  & Reconstruction task  & Backbone &  Pre-train  & Fine-tune Modality  & mIoU \\
\hline

\hline

\hline

MAE~\cite{mae}  & RGB                                                                      & ViT-B    &  ImageNet& RGB        & 47.0  \\ \hline

MAE~\cite{mae}  & RGB                                                                       & ViT-B      & ImageNet+SUN RGB-D  & RGB       &  47.3 \\ \hline

Mask3D~\cite{mask3d}   & Depth                                                                       & ViT-B    & ImageNet+SUN RGB-D       & RGB        & 47.4  \\ \hline

MultiMAE~\cite{multimae}  & RGB + Depth                                                                       & ViT-B    & ImageNet+SUN RGB-D& RGB       & 47.1   \\ \hline

\textbf{Ours}  & Depth & ViT-B    & ImageNet+SUN RGB-D   & RGB        & \textbf{48.7}  \\
\hline

\hline

\hline

\end{tabular}
}%
\vspace{5pt}
\caption {\label{tab:sunrgbd} Comparison with the state-of-the-art approaches on \textbf{SUN RGB-D Semantic Segmentation}. }
\end{table*}
\begin{table*}[h!]
\centering

\resizebox{0.99\textwidth}{!}{%
\begin{tabular}{c|c|c|c|c|c}
\hline

\hline

\hline
Methods  & Reconstruction task  & Backbone &  Pre-train  & Fine-tune Modality  & $\delta_{1}$ \\
\hline

\hline

\hline

MAE~\cite{mae}  & RGB                                                                      & ViT-B    &  ImageNet& RGB        & 85.1  \\ \hline

Mask3D~\cite{mask3d}   & Depth                                                                       & ViT-B    & ImageNet+ScanNet       & RGB        & 85.4  \\ \hline

CroCo~\cite{croco}  & RGB + Depth                                                                       & ViT-B    & Habitat& RGB       & 85.6  \\ \hline

MultiMAE*~\cite{multimae}  & RGB + Depth + Segmentation                                                                      & ViT-B    & ImageNet & RGB       & 83.0  \\ \hline

MultiMAE~\cite{multimae}  & RGB + Depth                                                                       & ViT-B    & ImageNet+ScanNet& RGB       & 85.3  \\ \hline

\textbf{Ours}  & Depth & ViT-B    & ImageNet+ScanNet   & RGB        & \textbf{87.1}  \\
\hline

\hline

\hline
MultiMAE~\cite{multimae} & RGB + Depth + Segmentation & ViT-B & ImageNet  & RGB       & \textbf{86.4} \\ \hline
\end{tabular}
}%
\vspace{5pt}
\caption {\label{tab:nyu2_depth} \textbf{NYUv2 Depth Estimation}. Ours approach outperforms Mask3D and MultiMAE which shows that learned representations can be transferred to the dense prediction tasks even for different dataset.* means the number are reported from CroCo~\cite{croco} paper.}
\end{table*}
\begin{table}[h!]


\resizebox{0.43\textwidth}{!}{%
\begin{tabular}{c|c|c}
\hline

\hline

\hline
Methods  &  Pre-train  & \textit{acc$_{s}$} \\
\hline

\hline

\hline

CoMAE~\cite{comae}   & SUN RGB-D   & 64.3 \\ \hline

\textbf{Ours}   & ImageNet + SUN RGB-D    & \textbf{65.4}  \\
\hline

\hline

\hline

\end{tabular}
}%
\caption {\label{tab:sunrgbd_acc} Comparison with CoMAE~\cite{comae} using average accuracy (\textit{acc$_{s}$}) on \textbf{SUN RGB-D Semantic Segmentation}. }
\end{table}

\subsection{Depth Estimation}
In this section, we study the efficacy of our pre-trained model for depth estimation task. We use NYUv2~\cite{nyuv2} for this downstream task and report $\delta_{1}$ on the test set which shows the percentage of pixels that have an error ratio ($\max\{ \frac{\hat{y}_{p}}{y_{p}}, \frac{y_{p}}{\hat{y}_{p}} \}$) below 1.25~\cite{delta1_eval}. We use DPT~\cite{DPT} as dense prediction head following MultiMAE for all the approaches and use the model pre-trained on ScanNet dataset for fine-tuning. The results on the depth prediction task are reported in Table~\ref{tab:nyu2_depth}. We can draw the same observation from the table that our approach outperforms state-of-the-art approaches. Specifically, it achieves 87.1 $\delta_{1}$ compared to 85.4 $\delta_{1}$ for Mask3D and 85.3 $\delta_{1}$ for MultiMAE. We also compare our approach with CroCo~\cite{croco}, a cross-view completion pre-training approach specifically designed for 3D vision tasks. In particular, we achieve an improvement of \textbf{+1.5} $\delta_{1}$ over CroCo. These results again verify the generalizibility of our two-stage pre-training strategy by transferring the representations to various downstream tasks and datasets.


















\begin{table}[h!]


\resizebox{0.43\textwidth}{!}{%
\begin{tabular}{c|c|c}
\hline

\hline

\hline
Methods  &  Pre-train  & AP \\
\hline

\hline

\hline

Scratch      & None             &  12.2 \\ \hline

ImageNet baseline    & Supervised ImageNet   & 17.6 \\ \hline

Pri3D~\cite{pri3d_}    & ImageNet + ScanNet      & 18.3 \\ \hline

MoCov2~\cite{mocov2}    & ImageNet + ScanNet         & 18.3 \\ \hline

MAE~\cite{mae}  & ImageNet + ScanNet  & 20.7 \\ \hline

MultiMAE~\cite{multimae} & ImageNet + ScanNet   & 22.4  \\ \hline

Mask3D~\cite{mask3d}   & ImageNet + ScanNet   & 22.8 \\ \hline

\textbf{Ours}   & ImageNet + ScanNet   & \textbf{23.7}  \\
\hline

\hline

\hline

\end{tabular}
}%
\caption {\label{tab:2dinstance} Comparison with other state-of-the-art approaches using average precision (AP) on \textbf{ScanNet 2D Instance Segmentation}. }
\end{table}
\subsection{Instance Segmentation}
We also demonstrate the effectiveness of our approach on 2D instance segmentation. We use ScanNet~\cite{scannet} for comparison against other baselines and use MaskRCNN~\cite{maskrcnn} as the instance segmentation head on top of the ViT-B encoder. Table~\ref{tab:2dinstance} shows the results on 2D instance segmentation. We see the same observation that our approach transfers the learned representations better by outperforming the state-of-the-art approaches including Mask3D and MultiMAE.



\subsection{Ablation Studies}


\begin{table}[h!]

\centering

\begin{tabular}{|c|c|c|}
\hline

\hline

\hline
Method  & Backbone  & mIoU \\
\hline

\hline

\hline

w/out noise & ViT-B & 66.5 \\ \hline

noise only  & ViT-B & 66.9 \\ \hline

Full & ViT-B & \textbf{67.5} \\ \hline

\end{tabular}
\caption {\label{tab:ablation_denoising} Ablation study on the components of the denoising method. We report the performance on \textbf{ScanNet 2D semantic segmentation}.}
\end{table}
\paragraph{Denoising.} 
We consider three settings in this ablation study: 1) Without any noise 2) Input with noise only 3) full method with all the denoising components. Table~\ref{tab:ablation_denoising} shows the performance of the different components of denoising loss on ScanNet 2D segmentation. It can be seen from the table that by simply adding noise in the input, the performance improves 0.4 mIoU. It is further improved by 0.6 mIoU with full denoising method that consists of noise prediction and positional encoding in the decoder. 


\begin{table}[tp]
    \centering
    \setlength\tabcolsep{4.5pt}
        \begin{tabular}{cccc|c}
        \textbf{Contrastive} & \textbf{Reconstruction} & \textbf{Denoising} & \textbf{Distill} & \textbf{mIoU} \\
        \hline
    
        \hline
    
        \hline
        \cmark & \xmark & \xmark & \xmark &  63.4 \\
        \hline
        \cmark & \cmark & \xmark & \xmark & 66.3 \\
        \cmark & \cmark & \xmark & \cmark & 66.5 \\
        \cmark & \cmark & \cmark & \xmark &  67.0 \\
        \hline
        \cmark & \cmark & \cmark & \cmark &  \textbf{67.5}\\ 
        \end{tabular}
    \caption{Effect of different pre-training objectives on \textbf{ScanNet 2D semantic segmentation}.}
    \label{tab:objective_mm}
\end{table}
\paragraph{Effect of pre-training objectives.}
In Table~\ref{tab:objective_mm}, we highlight the importance of using different self-supervised frameworks. It shows that contrastive learning alone can perform reasonably well. By pre-training using only depth reconstruction loss in the second stage, the performance jumps from 63.4 to 66.3 which is on-par with Mask3D while integrating all three objectives in the stage-2 significantly enhances the final performance. Lastly, without our feature distillation, the performance drops from 67.5 to 67.0 mIoU.



\begin{table}[h!]

\centering

\begin{tabular}{|c|c|c|}
\hline

\hline

\hline
RGB ratio  & Depth ratio  & mIoU \\
\hline

\hline

\hline

20.0\% & 20.0\% & 65.6\\ \hline

20.0\% & 50.0\% & 65.2\\ \hline

20.0\% & 80.0\% & 65.5\\ \hline

50.0\% & 20.0\% & 65.9\\ \hline

50.0\% & 50.0\% & 65.4\\ \hline

80.0\% & 20.0\% & 65.9\\ \hline

80.0\% & 80.0\% & \textbf{67.5} \\ \hline

\end{tabular}
\vspace{5pt}
\caption {\label{tab:ablation_masking} Effect of different masking ratios for RGB and depth modality on \textbf{ScanNet 2D semantic segmentation}.}
\end{table}
\paragraph{Effect of masking ratio during pre-training.}
Table~\ref{tab:ablation_masking} shows the performance on ScanNet 2D semantic segmentation by varying the masking ratio of RGB and depth modalities. It can be seen that our approach achieves the best performance with higher masking ratio for both of the modalities, which accelerates the pre-training.

\begin{table}[h!]
\centering

\begin{tabular}{|c|c|c|}
\hline

\hline

\hline
Loss  & Backbone  & mIoU \\
\hline

\hline

\hline


Depth + RGB Reconstruction & ViT-B & 66.9 \\ \hline

Full & ViT-B & \textbf{67.5} \\ \hline

\end{tabular}
\caption {\label{tab:ablation_loss} Ablation study of RGB and Depth reconstruction in the second stage on \textbf{ScanNet 2D semantic segmentation}.}
\vspace{-10pt}
\end{table}
\paragraph{Effect of RGB + Depth Reconstruction.} We also study the effect of reconstructing the RGB input besides the depth modality. It can be seen from the table~\ref{tab:ablation_loss} that it doesn't improve the performance. A plausible explanation for this is that reconstructing the RGB modality during pre-training is relatively simpler, as the 3D priors are encoded through the depth prediction process, providing additional reinforcement.

\begin{table}[h!]
\centering

\begin{tabular}{|c|c|c|}
\hline

\hline

\hline
Method  & Backbone  & mIoU \\
\hline

\hline

\hline

MAE~\cite{mae} & ViT-B & 64.8 \\ \hline

MultiMAE~\cite{multimae} & ViT-B & 65.1\\ \hline

Ours & ViT-B & \textbf{67.5} \\
\hline

\hline

\hline

MAE~\cite{mae} & ViT-L & 68.2 \\ \hline

MultiMAE~\cite{multimae} & ViT-L & 69.3 \\ \hline

Ours & ViT-L & \textbf{70.8} \\ \hline

\end{tabular}
\caption {\label{tab:vit_variant} Performance on \textbf{ScanNet 2D semantic segmentation} using variants of ViT.}
\end{table}
\paragraph{Results on different ViT variants.}
In the main manuscript, we showed the comparison of different approaches using ViT-B architecture. In Table~\ref{tab:vit_variant}, we show the performance on ScanNet 2D segmentation using ViT-L. We can infer the same observation that our approach outperforms the baselines that include MAE and MultiMAE.



\vspace{5pt}
\subsection{Low-data Regime Scenarios} 
\begin{figure}[h!]
\centering
\includegraphics[width=0.53\textwidth]{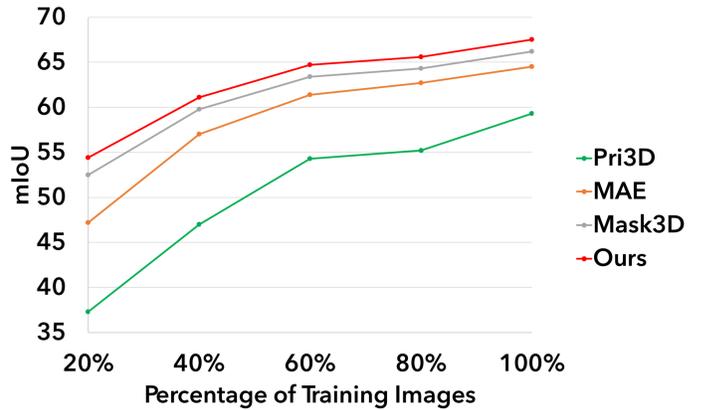}
\caption{To show data-efficiency feature of our approach, we compare with recent state-of-the-art pre-training approaches on \textbf{ScanNet 2D semantic segmentation} under limited labeled data scenarios.}
\label{fig:data_efficient}
\end{figure}

We show the data-efficient nature of our pre-training strategy by fine-tuning it in limited labeled data scenarios. Figure~\ref{fig:data_efficient} shows the performance on ScanNet 2D semantic segmentation. Our approach consistently outperform Mask3D, MAE and Pri3D across different ranges of limited labeled data. More notably, it marginally outperforms MAE with only 60\% of the original training set. 

\section{Conclusion}
We propose a new progressive pre-training strategy that effectively combines powerful self-supervised learning methods in a single framework to pre-train ViTs using RGB-D datasets. Our pre-training consists of two phases of learning: (1) contrastive learning which aligns the RGB and depth patches to extract discriminative local context features, (2) Masked Image Modeling which reconstructs the masked patches in the depth modality using RGB-D input to learn spatial statistical correlations and embed 3D priors. To learn high frequency features, we introduce a denoising loss inspired by the recent success of diffusion models to predict the noise added in the input. To further leverage the knowledge learnt in the stage-1, we employ feature distillation based on the global embeddings. We demonstrate the effectiveness of our approach on various downstream tasks such as semantic segmentation, depth prediction and instance segmentation. Finally, we show scalability and data-efficiency by pre-training the second stage using small scale RGB-D datasets.


{\small
\bibliographystyle{ieee_fullname}
\bibliography{references}
}

\clearpage
\setcounter{page}{1}

\begin{table*}[h!]
\centering

\resizebox{0.99\textwidth}{!}{%
\begin{tabular}{c|c|c|c|c|c}
\hline

\hline

\hline
Methods  & Reconstruction task  & Backbone &  Pre-train  & Fine-tune Modality  & mIoU \\
\hline

\hline

\hline

MAE~\cite{mae}  & RGB                                                                      & ViT-B    &  ImageNet& RGB        & 46.9  \\ \hline

MAE~\cite{mae}  & RGB                                                                       & ViT-B      & ImageNet+SUN RGB-D  & RGB       & 47.5  \\ \hline

Mask3D~\cite{mask3d}   & Depth                                                                       & ViT-B    & ImageNet+SUN RGB-D       & RGB        & 47.9  \\ \hline

MultiMAE~\cite{multimae}  & RGB + Depth                                                                       & ViT-B    & ImageNet+SUN RGB-D& RGB       & 47.4  \\ \hline

\textbf{Ours}  & Depth & ViT-B    & ImageNet+SUN RGB-D   & RGB        & \textbf{49.2}  \\
\hline

\hline

\hline

\end{tabular}
}%
\caption {\label{tab:nyu2} We report the mean IoU of ViT-B + ConvNeXt based segmentation header on \textbf{NYUv2 Semantic Segmentation}. }
\end{table*}
\section{More Results}

\subsection{NYUv2 Semantic Segmentation}
We show the generalizibility of the approach by fine-tuning the pre-trained model on NYUv2. More specifically, we use the model where the stage-2 pre-training is conducted using the SUN RGB-D dataset. It contains 1449 RGB-D images and we use the official split of 795 training images and 654 testing images. Table~\ref{tab:nyu2} shows the superiority of our approach compared to pre-training using Mask3D and other baselines for NYUv2 2D semantic segmentation.

\begin{table*}[h!]

    \centering
    
    \resizebox{0.55\textwidth}{!}{%
    \begin{tabular}{l|c}
    \hline
    
    \hline
    
    \hline
    Configuration & NYUv2~\cite{nyuv2} \\
    \hline
    
    \hline
    
    \hline
    Optimizer & {AdamW}\\
    Optimizer betas & {\{0.9, 0.999\}}\\
    Base learning rate & 1e-4\\
    Weight decay & 1e-4 \\
    Learning rate schedule & {cosine decay} \\
    Warmup epochs & 100\\
    Warmup learning rate & 1e-6 \\
    Epochs & 2000 \\
    Batch Size & 128 \\
    Layer-wise lr decay & 0.75\\
    \hline
    
    \hline
    Input resolution & 256 x 256 \\
    Augmentation & {RandomCrop, Color jitter} \\
    
    \hline
    
    \hline
    
    \hline
    \end{tabular}
    }%
    \caption{Fine-tune setting for NYUv2~\cite{nyuv2} \textbf{depth estimation}.}
    \label{tab:fine-tune_depth}
\end{table*}

\begin{table*}[h]
    \centering
    
    \resizebox{0.75\textwidth}{!}{%
    \begin{tabular}{l|cc}
    \hline
    
    \hline
    
    \hline
    Configuration & ScanNet~\cite{scannet} & SUN RGB-D~\cite{sunrgbd}\\
    \hline
    
    \hline
    
    \hline
    Optimizer & \multicolumn{2}{c}{AdamW}\\
    Optimizer betas & \multicolumn{2}{c}{\{0.9, 0.95\}}\\
    Base learning rate & \multicolumn{2}{c}{1e-4} \\
    Weight decay & \multicolumn{2}{c}{5e-2}  \\
    Learning rate schedule & \multicolumn{2}{c}{cosine decay} \\
    Stage-2 epochs & 100 & 250 \\
    \hline
    
    \hline
    Augmentation & \multicolumn{2}{c}{Gaussian Blur, ColorJitter} \\
    $\alpha$ & \multicolumn{2}{c}{1.0} \\
    $\beta$ & 0.01 & 0.1 \\
    $\gamma$ & \multicolumn{2}{c}{1.0} \\

    Masking ratio & \multicolumn{2}{c}{0.8} \\

    \hline
    
    \hline
    
    \hline
    \end{tabular}
    }%
    \vspace{5pt}
    \caption{Stage-2 pre-training setting on ScanNet~\cite{scannet} and SUN RGB-D~\cite{sunrgbd}.}
    \label{tab:pretrain_scannet_sunrgbd}
\end{table*}


\begin{table*}[t!]
    
    \centering
    
    \resizebox{0.75\textwidth}{!}{%
    \begin{tabular}{l|ccc}
    \hline
    
    \hline
    
    \hline
    Configuration & ScanNet~\cite{scannet} & NYUv2~\cite{nyuv2} & SUN RGB-D~\cite{sunrgbd}\\
    \hline
    
    \hline
    
    \hline
    Optimizer & \multicolumn{3}{c}{AdamW}\\
    Optimizer betas & \multicolumn{3}{c}{\{0.9, 0.999\}}\\
    Base learning rate & \multicolumn{3}{c}{1e-4}\\
    Layer-wise lr decay & \multicolumn{3}{c}{0.75}\\
    Weight decay & \multicolumn{3}{c}{5e-2}\\
    Learning rate schedule & \multicolumn{3}{c}{cosine decay} \\
    Warmup epochs & \multicolumn{3}{c}{1}\\
    Warmup learning rate & \multicolumn{3}{c}{1e-6} \\
    Drop path & \multicolumn{3}{c}{0.1} \\
    Epochs & 60 & 200 & 60\\
    \hline
    
    \hline
    Input resolution & 240 x 320 & 640 x 640 & 640 x 640 \\
    Color jitter & \xmark & \cmark &\xmark \\
    RandomGaussianBlur & \cmark  & \xmark & \cmark\\
    RandomHorizontalFlip & \cmark  & \xmark & \cmark\\    
    \hline
    
    \hline
    
    \hline
    \end{tabular}
    }%
    \vspace{5pt}
    \caption{Fine-tune setting on ScanNet~\cite{scannet}, NYUv2~\cite{nyuv2} and SUN RGB-D~\cite{sunrgbd} for \textbf{2D semantic segmentation}.}
    
    \label{tab:fine-tune_segmentation}
\end{table*}

\section{Implementation Details} \label{implementations}

\subsection{Pre-training and Fine-tuning Details}
In the stage-1, we pre-train the encoders for 300 epochs using ImageNet~\cite{imagenet} dataset using learning rate of 1.5e-4 with adamw optimizer. We report the stage-2 pre-training details on ScanNet~\cite{scannet} and SUN RGB-D~\cite{sunrgbd} in Table~\ref{tab:pretrain_scannet_sunrgbd}. Furthermore, we report the fine-tuning details for semantic segmentation task and depth estimation in Table~\ref{tab:fine-tune_segmentation} and Table~\ref{tab:fine-tune_depth} respectively. Finally, we follow Mask3D~\cite{mask3d} for instance segmentation and fine-tune the model using Detectron2 with 1x schedule.

\subsection{Model Architecture}
The pre-training model consists of modality-specific encoders and a shared decoder following MAE~\cite{mae}. The modality specific encoders are ViT-B while the decoder consists of 8 blocks with 16 multi-head attentions. The dimension is set to 512. We add a single fully connected layer on the top of the decoder for the depth reconstruction. For noise embedding, we use 2 Fully Connected (FC) layers and 1 ReLU. For fine-tuning, we follow MultiMAE~\cite{multimae} for downstream task head. For 2D semantic segmentation, we use ConvNeXt~\cite{convnext} based decoder while for depth estimation, we use DPT~\cite{DPT}. Lastly, we use MaskRCNN~\cite{maskrcnn} for instance segmentation.

\end{document}